\documentclass{article}

\usepackage{PRIMEarxiv}

\usepackage[utf8]{inputenc} 
\usepackage[T1]{fontenc}    
\usepackage{hyperref}       
\usepackage{url}            
\usepackage{booktabs}       
\usepackage{amsfonts}       
\usepackage{nicefrac}       
\usepackage{microtype}      
\usepackage{lipsum}
\usepackage{fancyhdr}       
\usepackage{graphicx}       
\graphicspath{{figures/}}     
\usepackage{multirow}
\title{Controllable Hybrid Captioner for Improved Long-form Video Understanding}

\author{
  Kuleen Sasse \textsuperscript{\textsection}, Efsun Sar{\i}o\u{g}lu Kay{\i} \textsuperscript{\textsection}, Arun Reddy \\
  {\tt\small Johns Hopkins University Applied Physics Laboratory}\\
{\tt\small Laurel, Maryland, USA}\\
  \texttt{\{kuleen.sasse, efsun.kayi, arun.reddy\}@jhuapl.edu} \\
}
\begin{document}
\maketitle
\begingroup\renewcommand\thefootnote{\textsection}
\footnotetext{Equal contribution}
\endgroup
\begin{abstract}
Video data, especially long-form video, is extremely dense and high-dimensional. Text-based summaries of video content offer a way to represent query-relevant content in a much more compact manner than raw video. In addition, textual representations are easily ingested by state-of-the-art large language models (LLMs), which enable reasoning over video content to answer complex natural language queries. To solve this issue, we rely on the progressive construction of a text-based memory by a video captioner operating on shorter chunks of the video, where spatio-temporal modeling is computationally feasible. We explore ways to improve the quality of the activity log comprised solely of short video captions. Because the video captions tend to be focused on human actions, and questions may pertain to other information in the scene, we seek to enrich the memory with static scene descriptions using Vision Language Models (VLMs). Our video understanding system relies on the LaViLa video captioner in combination with a LLM to answer questions about videos. We first explored different ways of partitioning the video into meaningful segments such that the textual descriptions more accurately reflect the structure of the video content. Furthermore, we incorporated static scene descriptions into the captioning pipeline using LLaVA VLM, resulting in a more detailed and complete caption log and expanding the space of questions that are answerable from the textual memory. Finally, we have successfully fine-tuned the LaViLa video captioner to produce both action and scene captions, significantly improving the efficiency of the captioning pipeline compared to using separate captioning models for the two tasks. Our model, controllable hybrid captioner, can alternate between different types of captions according to special input tokens that signals scene changes detected in the video. 

\end{abstract}

\section{Introduction}
\begin{figure*}[t]
  \centering
   \includegraphics[width=\linewidth]{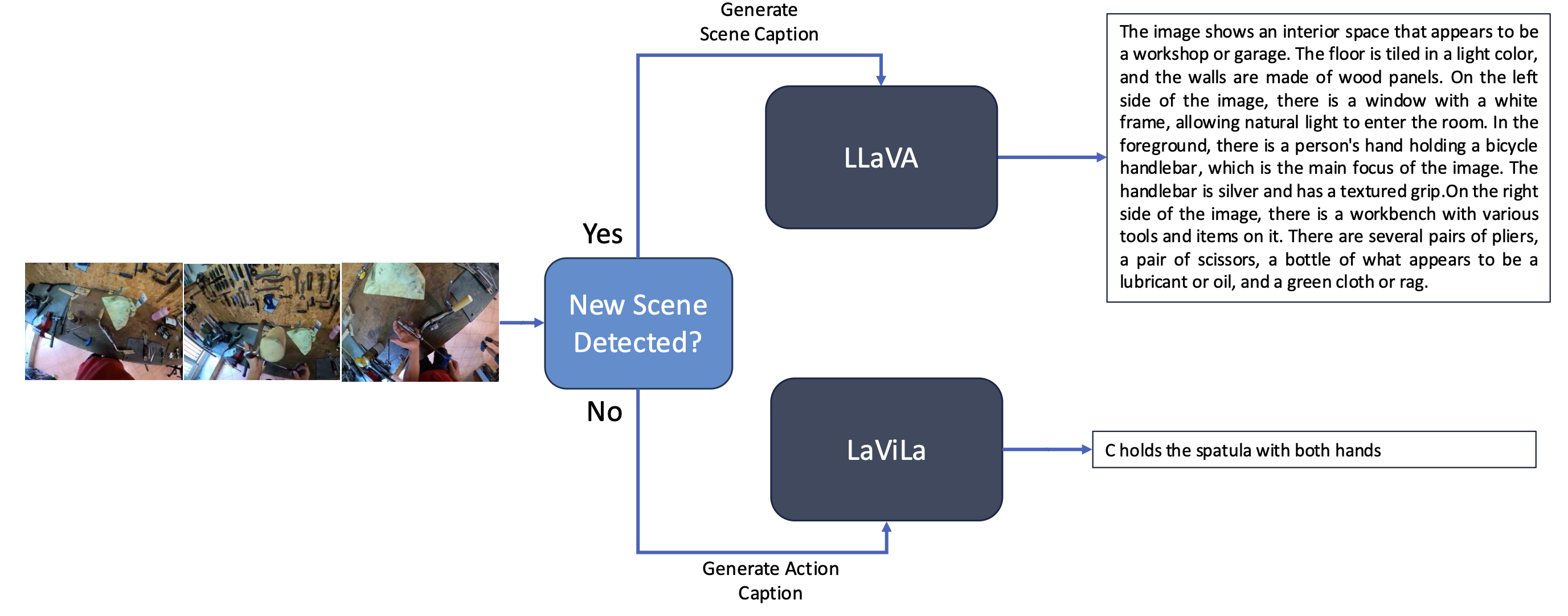}
   \caption{Ensemble Captioner: Two separate models are responsible for generating the action and scene captions.}
   \label{fig:ensemble}
\end{figure*}
\begin{figure*}[t]
  \centering
   \includegraphics[width=\linewidth]{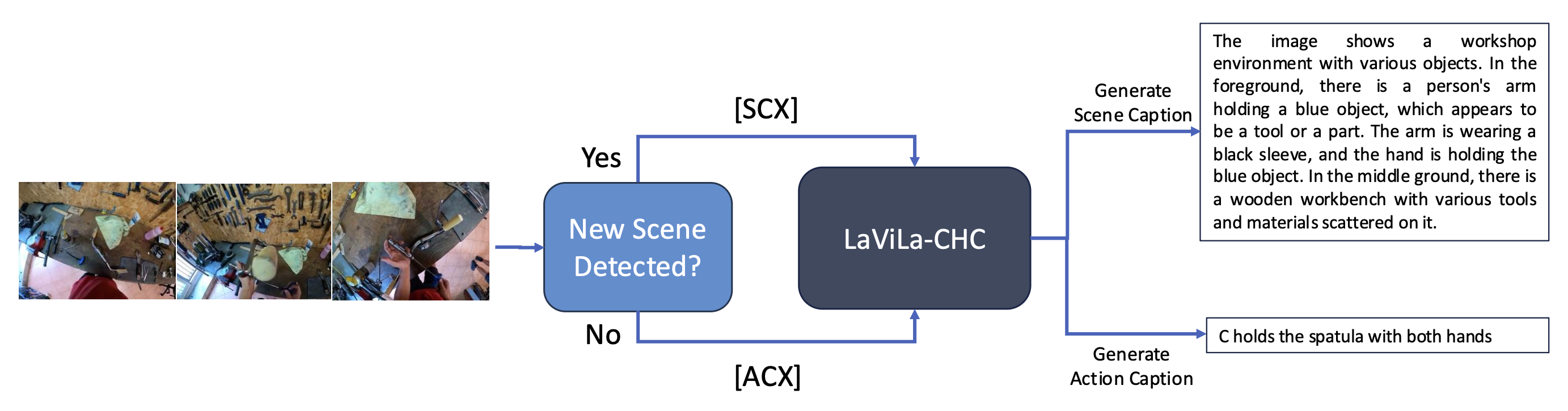}
   \caption{Controllable Hybrid Captioner (CHC): Single model generates both action and scene captions. }
   \label{fig:hybrid}
\end{figure*}
Current approaches for short-form video understanding do not easily scale up to processing and reasoning over longer videos. Video analysis research has largely focused on short-term tasks like human action recognition, which analyze video clips on the order of a few seconds in length \cite{7298698}. While useful, these algorithms do not scale easily to higher-level understanding of long-form videos, which involves reasoning over complex relationships between different actions, actors and objects \cite{10350674}. In addition, the spatiotemporal modeling present in short-form video models would become computationally infeasible to implement at longer timescales \cite{gberta_2021_ICML}. 

We investigate approaches for handling long-form video through the progressive construction of a compact \emph{memory} of observed activity. Our methods enable natural language question answering over several minutes of video data. To represent the video in a more compact format, we construct a text-based representation of the video content in the form of a log of different events that occur during the video. This log is then used to answer natural language questions about the video and evaluate our approaches on a publicly available video question answering benchmark, EgoSchema \cite{10.5555/3666122.3668126}. 

Specifically, our framework is based on Language-augmented Video Language Pretraining (LaViLa) \cite{10204456} captioner which learns video-language representations by conditioning LLMs on visual input, and finetunes them to create automatic video narrators.  The LLoVi \cite{zhang-etal-2024-simple} approach demonstrates how LaViLa model can be used as a short-term video captioner for long-form video Q\&A by pairing it with another LLM.  In contrast to LLoVi, we do not use multi-round
prompting or more advanced prompting strategies such as query-based summarization, and instead use the same prompt for Q\&A. Both LaViLa and LLoVi models focus only on action captions and do not capture other possibly relevant details such as scene information. In our approach, we improve long-form video Q\&A task by enhancing the caption logs to incorporate scene information when a scene change is detected. We first pair LaViLa captioner with a VLM to describe the scene. To improve efficiency, we also finetune the narrator on synthetically generated scene captions to act as a hybrid captioner. Our results show that having scene information in the caption log improves long-form video Q\&A accuracy. By incorporating scene information only when a change is detected, and training a single model to output both types of captions, we also improve the efficiency of the system. Our codebase for training and inferencing the controllable hybrid captioner has been open-sourced \footnote{https://github.com/jhuapl-fomo/lavila-chc}.

\begin{table*}
  \centering
  \begin{tabular}{@{}lllcc@{}}
    \toprule
   \textbf{Captioner Model}&	\textbf{Scene Segmentation} 	&\textbf{Scene Captions?}&	\multicolumn{2}{c}{\textbf{Accuracy (\%)}}\\
    \midrule
    LaViLa/LLoVi	&n/a	&No	&\multicolumn{2}{c}{41.4}\\
    \midrule 
    &	 	&&	  \multicolumn{2}{c}{\textbf{Scene Captioner }}\\
     \cline{4-5}
        &	 	&&	 \textbf{LLaVA 7B}& \textbf{LLaVA 34B}\\  
     \midrule
     \multirow{3}{*}{LaViLa + LLaVA} &	Uniform	& Yes &	38.2 &51.0\\
        &PyScene	& Yes &	37.6 &50.2\\
	 &	KTS	& Yes &	\textbf{43.8}&\textbf{57.2} \\
        \midrule
        &	 	&&  \multicolumn{2}{c}{\textbf{Distillation Model}}	 \\
        \cline{4-5}
        &	 	&&	 \textbf{LLaVA 7B}& \textbf{LLaVA 34B}\\ 
           \midrule
    \multirow{3}{*}{LaViLA-CHC}	
    &Uniform&	Yes	&\textbf{50.2} & \textbf{52.4}\\ 
    &PyScene&	Yes&	48.4&51.6\\	 
	&KTS	&Yes	&40.6 & 52.2\\
    \bottomrule
  \end{tabular}
  \caption{Accuracy of our captioning framework on EgoSchema dataset using Llama3.1-70B-Instruct for Q\&A }
  \label{tab:egoschema-controlled-ensemble}
\end{table*}

\section{Methodology}
\label{sec:methodology}
For better long-form video understanding and Q\&A, we improve action-only caption logs generated by LaViLa and LLoVi models by incorporating scene information. To avoid data leakage, we use LLoVi-base model that is trained by removing EgoSchema videos from Ego4D \cite{Grauman_2022_CVPR} pretraining data. 
For Q\&A reasoning backbone, we used the open-source Llama3.1-70B-Instruct \cite{Dubey2024TheL3} model to allow reproducibility of our results.

\begin{table*}
  \centering
  \begin{tabular}{lcccccc}
  \toprule
\textbf{Caption Type}&\textbf{Captioner Model}&\textbf{Teacher Model}&\textbf{BLEU}&\textbf{ROUGE}&\textbf{METEOR}&\textbf{Sentence BERT}\\
\midrule
\multirow{3}{*}{Action}	&LaViLa&None&	13.41	& 0.39 & 	0.36&	0.49 \\
&	LaViLa-CHC & LLaVA 7B &13.71	&0.37&	0.32&	0.66 \\ 
&	LaViLa-CHC & LLaVA 34B &	
13.71	& 0.38	&0.32	&0.66 \\
\midrule
\multirow{2}{*}{Scene}	&LaViLa-CHC & LLaVA 7B	&
1.56	&0.10&	0.17&	0.63\\
&	LaViLa-CHC & LLaVA 34B	
&2.40&	0.10	&0.19&	0.68 \\
\midrule
\multirow{2}{*}{Overall}	&LaViLa-CHC & LLaVA 7B	&
7.50&	0.23&	0.24&	0.64\\
&	LaViLa-CHC & LLaVA 34B	&
7.93&	0.24&	0.25&	0.67\\
    \bottomrule
  \end{tabular}
  \caption{Lexical and semantic similarity scores between \emph{LaViLa-CHC} and ground-truth action captions and synthetic scene captions}
  \label{tab:lex_sim}
\end{table*}

\subsection{Ensemble Video Captioner}
The first approach for enriching video caption logs is an ensemble system where the LaViLa captioner is paired with the LLaVA VLM \cite{liu2023llava,liu2023improvedllava} as shown in Figure \ref{fig:ensemble}. When a scene change is detected, LLaVA VLM is asked to describe the scene with the prompt \emph{"Describe the scene as specifically as possible focusing on objects and their properties and their relations to other objects in the scene. Be as concise as possible like you are writing a log."}. We considered two sizes of LLaVA, 7B \footnote{llava-hf/llava-v1.6-vicuna-7b-hf}and 34B \footnote{llava-hf/llava-v1.6-vicuna-34b-hf}.

\subsection{Controllable Hybrid Video Captioner}
The second approach, called Controllable Hybrid Captioner LaViLa-CHC, finetunes the LaViLa narrator to add the capability of generating both action and scene captions as shown in Figure \ref{fig:hybrid}. For training data, we randomly sampled 350 Ego4D videos excluding EgoSchema videos and chose the center frame among 32 uniformly sampled frames. We then captioned each frame with the LLaVA model resulting in a dataset of size of about 200k. To condition LaViLa model's LLM GPT-2 \cite{radford2019language} on the type of caption to generate, we added two new tokens $[ACX]$ and $[SCX]$ to its vocabulary and paired action captions with $[ACX]$ and synthetic scene captions with $[SCX]$ inspired by \cite{keskarCTRL2019}. Once trained in this fashion, these tokens can then be used to prompt the captioner to output specific kind of captions. Compared to the ensemble captioner that includes LLaVA 7B or 34B as VLM, LaViLa's LLM GPT-2-medium has only 137 million parameters and thus improves the overall efficiency of the system. On the other hand, GPT-2 is a causal language model with no additional post-training such as instruction tuning or reinforcement learning from human feedback \cite{10.5555/3600270.3602281}. Accordingly, it can be susceptible to repetition in longer generations. When we extended the generation from single-sentence action descriptions to paragraph-long scene descriptions, the captioner occasionally generated repetitive content. To resolve this, we applied repetition penalty \cite{keskarCTRL2019}. We explored several values ($1, 1.2, 1.5, 2$ and $3$) and picked the best according to the Q\&A performance. A repetition penalty of $3$ worked the best for most setups. 

\subsection{Temporal Segmentation}
To decide when to add additional scene information, we explored several scene segmentation methods, namely uniform sampling, PySceneDetect\footnote{https://github.com/Breakthrough/PySceneDetect} and Kernel Temporal Segmentation (KTS)\cite{Afham_2023_ICCV}. We used the \emph{ContentDetector} algorithm of \emph{PySceneDetect}, which detects changes in color and intensity between adjacent frames using a weighted average of pixel changes and compares it against a set threshold to trigger a scene cut.

KTS, on the other hand, combines spatio-temporal video segmentation and region tracking to extract important semantic objects from videos based on multiple cues (such as color, edges, and motions), and kernel-based models. We used the ViT-B/32 CLIP model \cite{pmlr-v139-radford21a} to extract features of the sampled frames. Finally, we also considered that the scene changes happen uniformly at every 120 seconds. On average, \emph{PySceneDetect} detected 30 scenes compared to 20 for KTS.

\section{Experiments}
\label{sec:experiments}
We experimented with egocentric long-form video dataset EgoSchema \cite{10.5555/3666122.3668126} which is based on the Ego4D dataset.  
EgoSchema consists of over 5,000 multiple choice question answer pairs over 250 hours of video covering a range of natural human activity and behavior. For each question, it requires the correct answer to be selected between five options based on a three-minute-long video clip. As an evaluation metric, we report Q\&A accuracy i.e., the percentage of correctly answered questions.

EgoSchema accuracy results are shown in Table \ref{tab:egoschema-controlled-ensemble}. In contrast to LLoVi, we have not explored multi-round prompting or more advanced prompting strategies for Q\&A and opted for using a single prompt. LLoVi's standard prompting approach with GPT3.5 for Q\&A achieved $42.8$ accuracy whereas ours with Llama-3.1-70B-Instruct is $41.4$. We improved this baseline score by incorporating scene information to the action log. Having additional scene information in the video caption log improved the accuracy of the system for almost all experimental setups. Among the temporal segmentation approaches, uniform sampling achieved the best score for LaViLa-CHC model, whereas KTS approach was the best for LaViLA+LLaVA ensemble system. LLaVA-34B as teacher or as part of the system produced better scores than LLaVA-7B. LaViLA-CHC model performed better than the ensemble LaViLA+LLaVA on 7B setup while requiring a much smaller memory.

In addition to evaluating our framework's Q\&A performance on the EgoSchema dataset, we also computed lexical and semantic similarity measures between LaViLa-CHC captions and ground-truth action captions and LLaVA synthetic scene captions. For this experiment, we randomly sampled 50 videos from the Ego4D dataset excluding EgoSchema videos, which resulted in about $~100,000$ captions. For action captions, the similarity was calculated between ground-truth action captions and LaViLa-CHC generated action captions. For scene, LaViLa-CHC captions are compared against synthetic LLaVA captions. We report BLEU-4 \cite{10.3115/1073083.1073135}, ROUGE-L \cite{lin-2004-rouge} and METEOR \cite{banarjee2005} for lexical similarity and SentenceBERT\cite{reimers-2019-sentence-bert} \emph{all-mpnet-base-v2} model for semantic similarity. The results are shown in Table \ref{tab:lex_sim}. The action captions generated by LaViLa-CFC have higher semantic similarity scores than the base model. This improvement could be due to additional training that the model has gone through to learn scene captions. On the other hand, the scene captions have lower lexical similarity scores compared to the action captions while having similar semantic similarity. Since, the scene captions are longer than the action captions, it is likely to have more lexical variety which can contribute to lower lexical similarity. Finally, between the two teacher models with different sizes, there is a slight improvement of using LLaVA-34B model over LLaVA-7B model.

\section{Conclusion}
In this work, we present a new long-form video captioner framework which detects scene changes and alternates between action and scene captions accordingly. We show its effectiveness on long-form video Q\&A benchmark EgoSchema. While the straightforward approach of having several models with different purposes \emph{e.g.} action and scene, can work, we also show that by finetuning the video captioner, we can achieve better performance while hosting only a single model. Furthermore, by introducing special tokens to explicitly refer to the kind of caption to generate, we demonstrate controllable generation capability can be extended to incorporate other information such as outputs from object detection or semantic segmentation systems. For future work, we plan to explore smaller LLMs for Q\&A by enriching the caption log with relevant details. We believe that a rich log could be a way for all questions can be answered with with less capable LLMs, enabling smaller memory footprints for video caption and Q\&A system.

\bibliographystyle{unsrt}  
\bibliography{references}  

\end{document}